\documentclass[sigplan,screen]{acmart}
\AtBeginDocument{%
  \providecommand\BibTeX{{%
    \normalfont B\kern-0.5em{\scshape i\kern-0.25em b}\kern-0.8em\TeX}}}

\copyrightyear{2023} 
\acmYear{2023} 
\setcopyright{acmlicensed}\acmConference[BCB '23]{14th ACM International Conference on Bioinformatics, Computational Biology and Health Informatics}{September 3--6, 2023}{Houston, TX, USA}
\acmBooktitle{14th ACM International Conference on Bioinformatics, Computational Biology and Health Informatics (BCB '23), September 3--6, 2023, Houston, TX, USA}
\acmPrice{15.00}
\acmDOI{10.1145/3584371.3613002}
\acmISBN{979-8-4007-0126-9/23/09}

\begin{document}

\title{Can Attention Be Used to Explain EHR-Based Mortality Prediction Tasks: A Case Study on Hemorrhagic Stroke}

 \author{Qizhang Feng}
 \affiliation{%
   \institution{Texas A\&M University}
   \streetaddress{xxx}
   \city{College Station}
   \state{TX}
   \country{USA}
   \postcode{XXXXX}
}
\email{qf31@tamu.edu}

 \author{Jiayi Yuan}
 \affiliation{%
   \institution{Rice University}
   \streetaddress{xxx}
   \city{Houston}
   \state{TX}
   \country{USA}
   \postcode{XXXXX}
 }
 \email{jy101@rice.edu}

 \author{Forhan Bin Emdad}
 \affiliation{%
   \institution{Florida State University}
   \streetaddress{142 Collegiate Loop}
   \city{Tallahassee, FL}
   \country{USA}}
 \email{femdad@fsu.edu}

 \author{Karim Hanna}
 \affiliation{%
   \institution{University of South Florida}
   \streetaddress{xxx}
   \city{Tampa}
   \state{FL}
   \country{USA}
   \postcode{XXXXX}
 }
 \email{khanna@usf.edu}

 \author{Xia Hu}
 \affiliation{%
   \institution{Rice University}
   \streetaddress{xxx}
   \city{Houston}
   \state{TX}
   \country{USA}
   \postcode{XXXXX}
 }
 \email{xia.hu@rice.edu}

 \author{Zhe He}
 \affiliation{%
   \institution{Florida State University}
   \streetaddress{142 Collegiate Loop}
   \city{Tallahassee, FL}
   \country{USA}}
 \email{zhe@fsu.edu}

\renewcommand{\shortauthors}{}

\begin{abstract}
  Stroke is a significant cause of mortality and morbidity, necessitating early predictive strategies to minimize risks. Traditional methods for evaluating patients, such as Acute Physiology and Chronic Health Evaluation (APACHE II, IV) and Simplified Acute Physiology Score III (SAPS III), have limited accuracy and interpretability. This paper proposes a novel approach: an interpretable, attention-based transformer model for early stroke mortality prediction. This model seeks to address the limitations of previous predictive models, providing both interpretability (providing clear, understandable explanations of the model) and fidelity (giving a truthful explanation of the model's dynamics from input to output). Furthermore, the study explores and compares fidelity and interpretability scores using Shapley values and attention-based scores to improve model explainability. The research objectives include designing an interpretable attention-based transformer model, evaluating its performance compared to existing models, and providing feature importance derived from the model.
\end{abstract}

\ccsdesc[500]{Computing methodologies~Neural networks}

\keywords{neural networks, explainability, transformers, attention}

\maketitle

\section{Introduction}

Stroke has been ranked as the fifth leading cause of death and more than 795,000 people have a stroke in the United States every year \cite {writing2009heart}. Moreover, stroke causes long-term disability which constrains individuals in daily functional activities such as hand movement, walking, and speech delivery \cite{hu2010early}. Among all strokes, 87\% are ischemic strokes and 13\% are hemorrhagic strokes. Raptures cause hemorrhagic stroke in a blood vessel in the brain. Early prediction of mortality can support decision-making and minimize the risk of death from hemorrhagic stroke \cite{jeng2008predictors}. 

Despite machine learning (ML) and deep learning (DL) approaches outperforming many statistical models for mortality prediction~\cite{payrovnaziri2020explainable}, clinicians still mostly prefer traditional risk scores such as Acute Physiology and Chronic Health Evaluation (APACHE II, IV), and Simplified Acute Physiology Score III (SAPS III). These severity scores are calculated based on linear functions using a limited number of risk factors and do not yield high sensitivity and specificity, \cite{knaus1991apache}. Even though ML- and DL-based approaches can yield high accuracy in predictions of diagnoses and outcomes, these techniques are often considered uninterpretable and potentially biased \cite{emdad2023towardsethics}. To gain sufficient trust from clinicians, researchers are building explainable AI (XAI) models for mortality predictions. Interpretability and fidelity are the two necessary components of explainability \cite{markus2021role}. Interpretability refers to the unambiguous, easily understandable but parsimonious explanation of the model, whereas fidelity refers to the accuracy of the explanation of the task model. Recently, even though many studies used Shapley values as the post hoc explainability method to provide global interpretation, they have not been used to explain models with longitudinal data and their fidelity is questionable~\cite{marcilio2020explanations}. 

To address the aforementioned research gaps, we propose a robust attention-based transformer model for early stroke mortality prediction while providing local interpretation of longitudinal data. In this study, we use 24-hour data from electronic health records (EHRs) in the intensive care setting to predict 2-7 day mortality of hemorrhagic stroke. We re-use the dataset in a recently published study \cite{emdad2023towards}, which included demographics, lab values, comorbidities, and vital signs of hemorrhagic stroke patients. Furthermore, this study explores and compares the fidelity of the proposed attention-based scores with other XAI methods.

\section{Related Work}

Harerimana et al. proposed a hierarchical attention network (HAN) for mortality prediction using MIMIC data \cite{harerimana2021deep}. HAN achieved 86\% accuracy and 87\% AUROC in in-hospital mortality prediction. However, this study did not provide a clear interpretability method for the proposed model. Heo et al. proposed an uncertainty awareness (UA) model with an uncertainty-awareness attention component for mortality prediction of sepsis and achieved an AUROC of 78\%, which outperformed deterministic or stochastic attention mechanisms~\cite{heo2018uncertainty}. Similar to Heo's study, Jun et al. implemented variational autoencoder (VAE) on UA attention framework to predict mortality using MIMIC III data with 79\% AUC and 32\% AUPRC \cite{jun2019stochastic}.

Regarding the explainability of the models, a recent systematic review on XAI for healthcare between 2011-2022 found the attention-based explainability method as an effective approach along with SHAP and LIME-based explanation approaches \cite{loh2022application}. Deng et al. implemented RNN, Gated Recurrent Unit (GRU), and LSTM for mortality, length of stay (LOS), and 30-day readmission prediction providing attention-based interpretability~\cite{deng2022explainable}. Studies indicate that SHAP and attention-based mechanisms provide an extensive explanation, but there is a need to compare these methods. Our study bridges a connection between SHAP values and attention scores to provide a better explanation using transformer-based multimodal models. 

\section{Method}
In this section, we first introduce the problem setting and the notation used. Subsequently, we elucidate the detailed design of our attention-based model.

\subsection{Preliminary}
In this study, we use two types of data, vital signs data as time series data and aggregated data (i.e., demographics,  lab \& chart values) as tabular data. Formally, we define the vital signs time series data as $V = {v_1, ..., v_N}$, the aggregated data as $A = {a_1, ..., a_N}$ and the prediction label as $M = {m_1, ..., m_N}$. $N$ is the number of the data sample. The $i$-th vital signs time series sample $v_i = [[v_{i,1,1}, ..., v_{i,1,24}], ..., [v_{i,7,1}, ..., v_{i,7,24}]]$ contains data collected for seven different channels over a 24-hour period, where $v_{i,t,c}$ indicates the $i$-th vital sign time series data collected from the $c$-th channel at the $t$-th time.
The $i$-th aggregated data contains 196 features as $a_i = [a_i^1, ..., a_i^{196}]$, where $a_i^f$ is the $f$-th feature from the $i$-th aggregated data.
All the data value are normalized, i.e. $v_{i,t,c}, a_i^f\in [0,1]$, and the task is a binary prediction where $m_i\in \{0, 1\}$.

\subsection{Model Architecture}

In order to manage data with various modalities, we propose a hierarchical model comprising two distinct components. The first component is a time series data processing module, responsible for processing vital signs collected at different time points. This module is built upon the transformer model, incorporating a temporal and spatial attention mechanism~\cite{grigsby2021long}. Consequently, attention scores can be determined for each channel at every hour of the vital sign.

The transformer model consists of an encoder and a decoder. The encoder embeds time series data into a latent vector, represented as $E(v_{i,t,c}) \in \mathbb{R}^d$. The decoder is capable of converting the latent vector back into the original time series data space, supporting the self-supervised learning task. This will be further discussed in Section~\ref{sec:train_approach}.

\begin{figure}[h!]
\centering
\includegraphics[width=\linewidth]{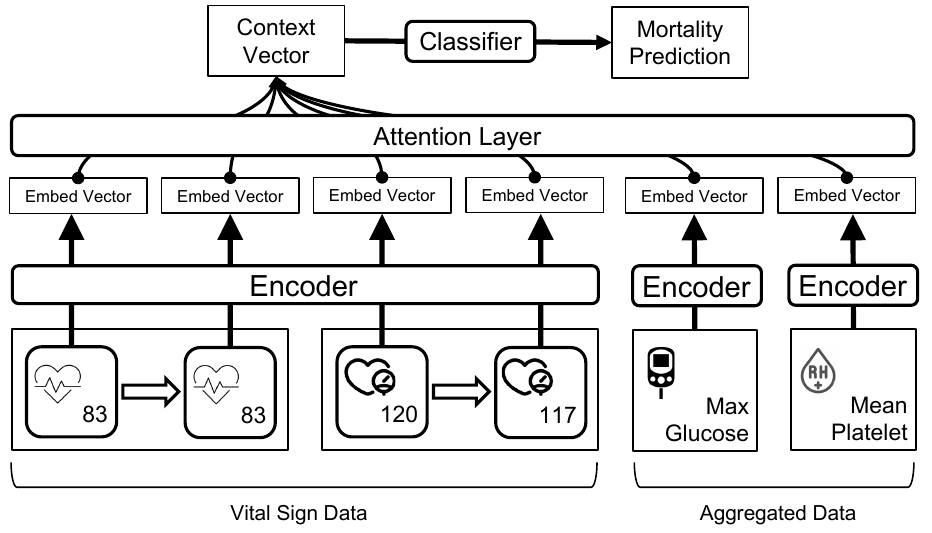}
\vspace{-2em}
\caption{Stage two mortality prediction training framework, integrating vital sign time series data and aggregated data. The pre-trained encoder from stage one is shared among vital sign data, while each aggregated data feature has its own dedicated encoder module. The attention layer fuses the embedded vectors of vital sign data and aggregated data into a context vector, used for mortality prediction.}
\vspace{-1em}
\label{fig:mort_modle}
\end{figure}

The second component, the mortality prediction module is a self-attention module that processes both the encoded time series data and the aggregated data as shown in Figure~\ref{fig:mort_modle}. In order to align the aggregated data with the latent space of the encoded time series data, we employ a two-layer MLP (multi-layer perceptron) with LeakyReLU as the activation function. This embeds each aggregated data feature $A$ into the same latent space as the embedded time series data, represented as $\text{MLP}(a_i^f) \in \mathbb{R}^d$. After embedding all the vital sign data and aggregated data into the latent space, a self-attention layer is applied to combine them into a prediction vector. Ultimately, a feed-forward layer outputs the probability for mortality prediction.

\subsection{Training Approach}
\label{sec:train_approach}
To train the hierarchical model, we devise a two-stage training strategy. The initial stage encompasses a self-supervised training process for the vital sign data processing module. To address the challenges posed by the heterogeneous nature of time series data and the limited availability of training data, we design a training objective that includes two tasks: a predictive task and a reconstruction task.
To be specific, we split the $i$-th vital sign data for each channel into past time data $[v_{i,t-p,c},..., v_{i,t,c}]$ and future time data $[v_{i,t+1,c},..., v_{i,t+q,c}]$.
The self-supervised training comprises two tasks. The first task involves utilizing past time data to predict future time data, while the second task aims to reconstruct the vital sign data from the embedding vector. Specifically, we mask the future time data values with 0 and input both past and masked future vital sign data into the vital sign data processing module. The encoder embeds the vital sign data into an embedded vector, and then the decoder generates predictions for the future data in the future vital sign data prediction task and reconstructs the past time vital sign data for the reconstruction task.

In our experiments, the past time data length is 12 hours, and the future time data length is 8 hours. For instance, we employ vital sign data collected from hour 1 to hour 12 to predict the vital sign data gathered from hour 13 to hour 20 and progressively shift the starting and ending moments to hour 5 and hour 24. This approach increases the training volume by five times the amount of data. We utilize mean square error as the loss function.

The second stage is supervised training for the mortality prediction task. We reuse the encoder of the vital sign processing module, pre-trained in the first stage, to embed the vital signal data. For each feature of aggregated data $a_i^f$, we train a new encoder for embedding. After generating the context vector via the attention layer, a classifier is applied to predict the mortality probability, and we employ cross-entropy loss as the loss function.

\subsection{Attention score as explanation}

The core component of our model is the attention mechanism that guides the model's focus toward relevant tokens during the learning phase. The mechanism operates by using two tokens, represented as $x$ and $z$. It is important to denote that the token $z$ is mapped onto key vectors using learned parameters, referred to as $W^k$. Concurrently, $x$ is engaged to generate corresponding query vectors through $W^q$.

Diving deeper into the mechanics of the attention model, there exist different types of attention expressions, such as scaled dot-product attention and additive attention. The former, which we are using, normalizes the dot product of query and key vectors with the square root of the dimension $d$, as demonstrated in the following equation:

\vspace{-1em}
\begin{equation}
\text{Attention}(x,z) = \text{softmax}\left( \frac{W^q(x)(W^k(z))^T}{\sqrt{d}}\right).
\end{equation}

In this context, the attention score functions as a metric of the weight ascribed to a specific token. The efficacy of this score is evidenced by its comprehensive implementation in a broad range of fields. It has been applied effectively in graph neural networks, natural language processing, and computer vision~\cite{velivckovic2017graph}. The adaptability of the attention mechanism is further highlighted by its utility as an intrinsic explanatory method, as emphasized in the study by Liu et al. (2022)~\cite{liu2022interpretability}.

In our research, the attention score, extracted from the self-attention module, serves as the explanation method. Thus, we introduce an element of interpretability into the model: features attributed with higher attention scores are considered more influential for the prediction task. This approach aids in better comprehending the model's predictions, enabling the evaluation of the impact and relevance of each feature in the model's decision-making process.

\section{Experiments}

In this section, we first introduce the data employed in our experimental procedures. Then we provide the baseline methods, along with the evaluation metrics. Finally, we show the results of the main experiment, both in terms of quantitative measurements and qualitative assessments.

\subsection{Dataset}

\begin{figure}[H]
  \centering
\includegraphics[width=1\linewidth]{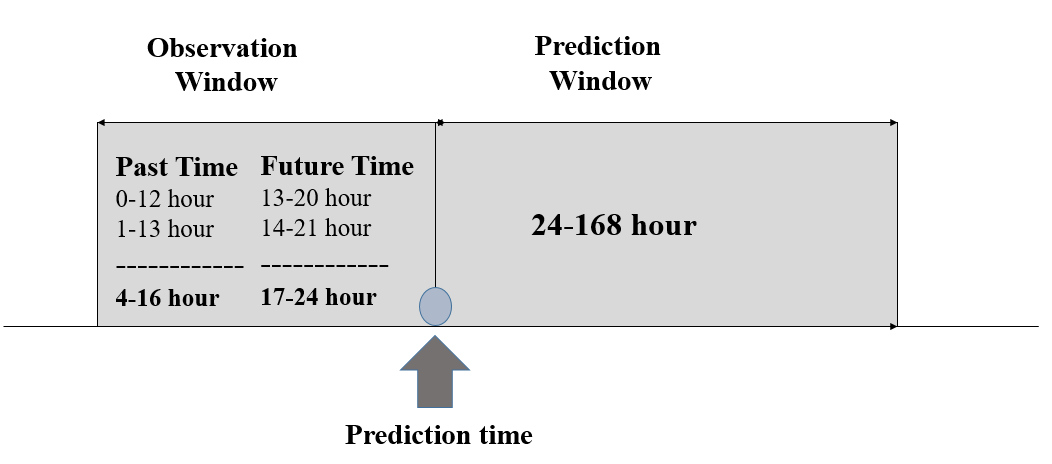}
 \caption{Observation and prediction windows}
 \label{fig:prediction_windows}
\end{figure}

In this study, we used the same dataset in a recently published study ~\cite{emdad2023towards} which extracted data from Medical Information Mart for Intensive Care (MIMIC)-III database. Variables include demographics, comorbidities, lab results, and vital signs. The time frame based on which mortality prediction is performed is known as the observation window and the time frame on which actual prediction is made is called the prediction window. In our study, our observation window was 24 hours and the prediction window was between 24 to 168 hours (2-7 days) (see Figure~\ref{fig:prediction_windows}). Initially, the dataset included a total of 2187 ICU stays after sorting the data with ICD-9 CM codes ('430', '431', '436', '432.0', '432.1', and '432.9') related to hemorrhagic stroke. After removing the inconsistent data, the total data remained of 2089 ICU stays. This dataset was highly imbalanced as the mortality rate was only 14\%. To reduce the imbalance ratio, we undersampled the negative instances. After undersampling, the total number of ICU stays became 614, where the ratio of positive and negative instances was 50-50\%. We used the MICE imputation technique based on multivariate imputation by chained equations for imputing the missing values in the prepared data set. In addition, we normalized the data with a min-max scaler which scales the data ranging between 0 to 1.

This dataset included a total of 7 time series and 196 numerical features containing minimum, maximum, median, and mean values for each of the lab \& chart values, comorbidities, 3 demographic variables for age, gender, and race. The features were selected based on the clinical relevance identified from the literature and data availability. Time series data included vitals and aggregated data included demographics, comorbidities, and lab \& chart values. This dataset was prepared with the objective to assist in clinical decision-making and predicting health outcomes.

\subsection{Baselines}

\paragraph*{Model Baselines.} In our study, we use two baseline models. Firstly, we implement logistic regression~\cite{tolles2016logistic}, where all vital sign data is aggregated and used as the input. This model handles all features as one combined vector. Secondly, we introduce an LSTM-based~\cite{hochreiter1997long} model, known as the Fusion model. In this model, we utilize LSTM to handle time-series data while static data is processed by an MLP. After processing, both outputs are combined and used to generate the final model output. The LSTM-Fusion approach efficiently leverages temporal dependencies in our time-series data, while logistic regression provides a simple but robust baseline.

\paragraph*{Explanation Baselines.} We compare the explainability of our model with other intrinsic explanation methods as well as the post-hoc explanation method. We use the weight of the logistic regression model as its intrinsic explanation. The absolute values of the model weights corresponding to each feature represent the importance of the features. We also adopt the SHAP (SHapley Additive exPlanations)~\cite{NIPS2017_7062} as the explanation baseline. SHAP is a post-hoc method for explaining machine learning model predictions. It computes the contribution of each feature to the output. SHAP uses Shapley values from game theory to determine the feature contributions, generating an explanation for each prediction made by the model.

\subsection{Evaluation Metrics}

\paragraph*{Model and Explanation Metrics.}
We report the AUROC and AUPRC for performance evaluation of the models. AUROC measures the model's ability to distinguish between positive and negative classes, while AUPRC measures the trade-off between precision and recall.

To evaluate the quality of the model explanation without human-labeled ground truth, we follow the rule that the explanations should be faithful to the model and should identify important input features. We use fidelity as the metric, $\text{Fidelity}^+$ and $\text{Fidelity}^-$:
$\text{Fidelity}^+$ is defined as the difference in accuracy (or predicted probability) between the original predictions and the new predictions after masking out important input features; In contrast, $\text{Fidelity}^-$ studies prediction change by keeping important input features and removing unimportant features.

To provide more detail, our process starts by ordering the features based on their importance as determined by the explanation method. For $\text{Fidelity}^+$, we substitute the $k$ most important features with random values and calculate the decrease in AUROC or predicted probability. And for $\text{Fidelity}^-$, we substitute the $k$ least important features and calculate the decrease in AUROC (predicted probability). Therefore, the distinction between a good explanation under the $\text{Fidelity}^+$ and $\text{Fidelity}^-$ methods lies in the degree of decrease in the AUROC or predicted probability after substituting either the most crucial or the least essential features respectively. The larger the decrease when focusing on the most important features ($\text{Fidelity}^+$), the better the explanation, and conversely, the smaller the decrease when dealing with the least important features ($\text{Fidelity}^-$), the better the explanation.

\subsection{Result} 
\subsubsection*{Quantitative Result}

In this section, we present the quantitative results of our experiments. Table~\ref{tab:util_result} provides performance metrics for the Logistic, LSTM-Fusion, and Attention models. We used a 10-fold cross-validation approach in our experiments. The results indicate that the attention-based model outperforms the logistic and LSTM baselines in terms of both AUROC and AUPRC. Specifically, the attention model's AUROC of 0.8487 ($\scriptstyle\pm 2.75e-2$) outperforms the logistic model's score of 0.8215 ($\scriptstyle\pm 9.37e-3$) and the LSTM-Fusion model's score of 0.8170 ($\scriptstyle\pm 2.45e-3$), demonstrating superior classification performance and an increased ability to distinguish between positive and negative classes. In terms of AUPRC, the attention model also leads with a score of 0.9726 ($\scriptstyle\pm 9.65e-3$), compared to the logistic model's score of 0.9675 ($\scriptstyle\pm 2.31e-3$) and the LSTM-Fusion model's score of 0.9723 ($\scriptstyle\pm 1.42e-3$). This shows that the attention model is more effective in maintaining a high precision rate as recall increases.



\begin{table}[t]
\centering
\caption{Utility Performance Result}\label{tab:util_result}
\vspace{-1em}
\resizebox{\columnwidth}{!}{
\begin{tabular}{lccc}
\toprule
      & Logistic & LSTM-Fusion & Attention \\
\midrule
AUROC & 0.8215 $\scriptstyle\pm 9.37e-3$ & 0.8170 $\scriptstyle\pm 2.45e-3$ & \textbf{0.8487} $\scriptstyle\pm 2.75e-2$ \\
AUPRC & 0.9675 $\scriptstyle\pm 2.31e-3$ & 0.9723 $\scriptstyle\pm 1.42e-3$ & \textbf{0.9726} $\scriptstyle\pm 9.65e-3$ \\
\bottomrule
\end{tabular}
}
\end{table}


\begin{table}[h!]
\centering
\caption{Explanation and Efficiency Performance Result}\label{tab:exp_result}
\resizebox{\columnwidth}{!}{
\begin{tabular}{llccccc}
\toprule
          &       & \multicolumn{2}{c}{Logistic}    & LSTM-Fusion    & \multicolumn{2}{c}{Attention}    \\ \cmidrule(lr){3-4}\cmidrule(lr){5-5}\cmidrule(lr){6-7}
          &       & Weight   & SHAP & SHAP & Atten     & SHAP \\
\midrule
Fidelity+ & AUROC ($\downarrow$) & $-$0.07  & $-$0.22 & $-$0.25 & \textbf{$-$0.36} & $-$0.40 \\
          & AUPRC ($\downarrow$) & $-$0.07  & $-$0.05 & $-$0.06 & \textbf{$-$0.09} & $-$0.07 \\
          & Prob. ($\downarrow$)  & $-$0.24  & $-$0.40 & $-$0.24 & \textbf{$-$0.25} & $-$0.29 \\
\midrule
Fidelity- & AUROC ($\uparrow$) & $-$0.00  & $-$0.09 & $-$0.01 & \textbf{0.00}  & $-$0.01 \\
          & AUPRC ($\uparrow$) & $-$0.00  & $-$0.02 & $-$0.03 & \textbf{0.00}  & $-$0.03 \\
          & Prob. ($\uparrow$)  & $-$0.01  & $-$0.07 & $-$0.02 & \textbf{$-$0.00} & $-$0.07 \\
\bottomrule
\end{tabular}
}
\vspace{-0.5em}
\end{table}

To evaluate the explanation performance of the Logistic Regression, LSTM-Fusion, and Attention models, we use the absolute value of feature weights, the LSTM-Fusion model's SHAP score, and the Attention score as intrinsic explanations, respectively. We also employ SHAP as a post-hoc explanation method for each model. Table~\ref{tab:exp_result} summarizes our findings on the comparison of these explanation methods.

Our Attention model outperforms both the Logistic and LSTM-Fusion models in terms of $\text{Fidelity}^+$, which signifies the importance of the model's features, across all aspects - AUROC, AUPRC, and Probability. In terms of AUROC, where a larger decrease indicates better model explanations, our Attention model, with a score of $-$0.3617, shows a much more significant drop compared to both baseline models. The same trend is observed for AUPRC, where our model's scores are noticeably lower, implying that our model's explanations can indeed identify key input features better.

For $\text{Fidelity}^-$, a metric that evaluates the unimportance of the least contributing features, our Attention model again outperforms the Logistic and LSTM-Fusion models across all metrics. Our model demonstrates negligible changes in AUROC and AUPRC, and a smaller decrease in probability when the least important features are removed, which is desirable for Fidelity-. This confirms that our model's explanations accurately identify non-essential features.

Finally, it's worth highlighting that our Attention-based method also surpasses the SHAP-based method in terms of fidelity. As observed in Table~\ref{tab:exp_result}, for both $\text{Fidelity}^+$ and $\text{Fidelity}^-$ metrics, our Attention model consistently outperforms the SHAP-based methods associated with the Logistic and LSTM-Fusion models.

\begin{figure*}[h]
\centering
\includegraphics[width=.99\linewidth]{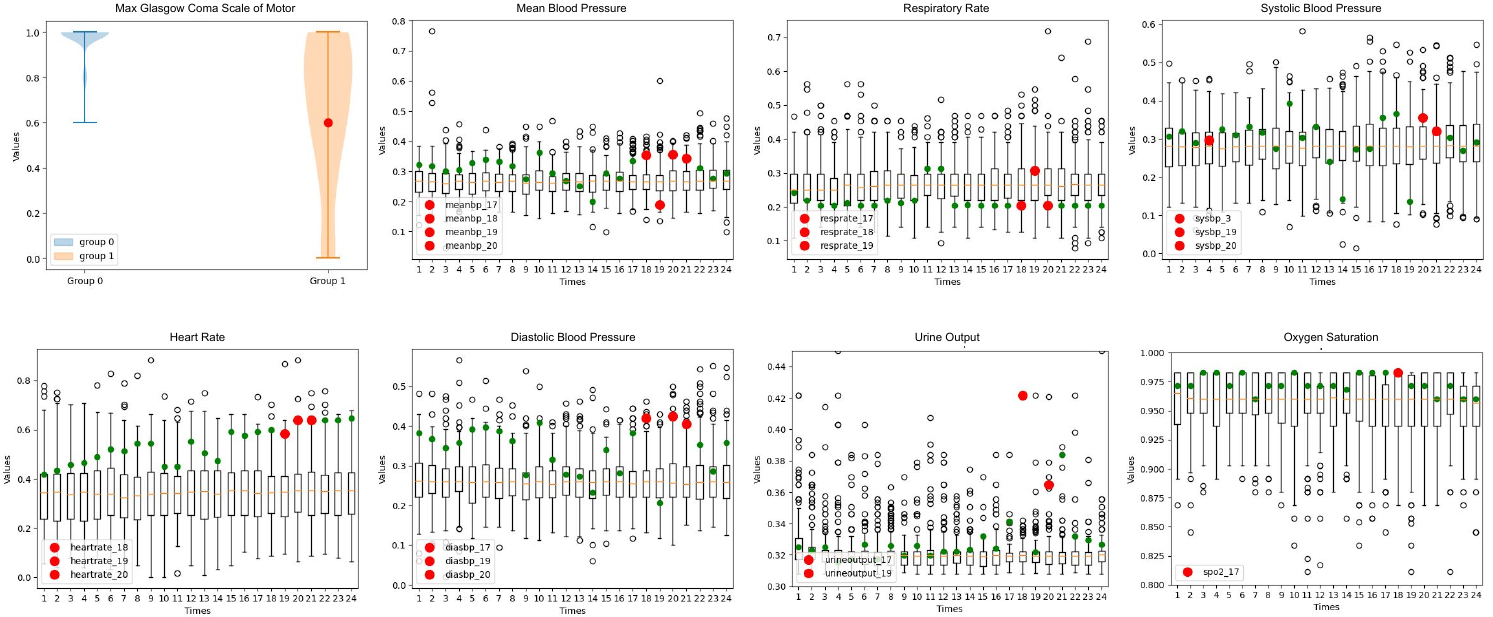}
\vspace{-5pt}
\caption{Visualization of the explanation of a patient whose mortality prediction is true.}

\label{fig:vis}
\end{figure*}

\subsubsection*{Qualitative Result}
In this section, we present a case study to demonstrate the qualitative result. Specifically, we visualize the explanation for a patient whose mortality prediction is positive (i.e., this patient died between 2-7 days in ICU). 
We use the attention score as the explanation, the feature with a higher attention score is more important to the prediction. We rank both the aggregated data attributes as well as the vital signs data attributes according to their attention score. The top 20 important attributes are picked for visualization.

If the important attribute from the explanation belongs to the aggregated data attribute, we plot the distribution of this attribute in the dataset as well as the value of the important attribute of the patient.
If the important attribute from the explanation belongs to the vital sign data attribute, we use the box plot to demonstrate the distribution of the vital sign data for each hour time. We plot the value of the attribute at each hour and mark the important ones in red.
If the attributes belong to the same vital sign, we plot them in the same figure.




Figure~\ref{fig:vis} shows the visualization of the important attributes for an explanation of a patient whose mortality prediction is true. From this figure, we can discern several key insights.
\begin{itemize}
    \item The important attributes are physiological indicators. For example, The Glasgow Coma Scale of motor (gcsmotor) is used to objectively describe the extent of impaired consciousness in all types of acute medical and trauma patients.
    \item Crucial attributes in vital sign data are usually located at a later time point, indicating that the patient's vital signs are in a state of deterioration in the final stage.
    \item The values of all important attributes are located outside the natural distribution, indicating that the patient is in an abnormal state at this point.
\end{itemize}

\section{Discussion and Conclusion}
This study introduced an attention-based transformer model for early mortality prediction in hemorrhagic stroke patients using longitudinal data from electronic health records (EHRs). The model's performance, interpretability, and fidelity were compared to traditional logistic regression models. Our results demonstrated the superior performance of the attention-based transformer model in mortality prediction, as measured by AUROC and AUPRC. Our model also offered more precise explanations for its predictions. The Fidelity scores showed that the attention-based model could identify key features better and accurately detect non-essential ones.

One of the key findings of this study was the observed improvement in performance when using the attention score as an intrinsic explanation, compared to the use of Shapley values. This demonstrates that the attention-based approach performs similarly to other post hoc explanation methods. The case study further highlighted the importance of features derived from the proposed attention-based transformer model, revealing that critical physiological indicators played a significant role in mortality prediction. Vital sign data attributes, especially those at later time points, were found to be particularly important, indicating that the patient's vital signs were deteriorating in the final stages. These attributes' values typically fell outside their natural distribution, confirming the patient's abnormal state.

In conclusion, the proposed attention-based transformer model has shown promise in early mortality prediction in hemorrhagic stroke patients, providing improved performance and interpretability over traditional logistic regression models. Future research should focus on validating these findings in other settings and disease states, and on enhancing the model's predictive power and explainability.

\section{Acknowledgement}

This study was partially supported by US National Library of Medicine grant R21LM013911 and US National Center for Advancing Translational Sciences grant 2UL1TR001427.

\bibliographystyle{ACM-Reference-Format}
\bibliography{sample-base}

\end{document}